\documentclass[graybox]{svmult}

\usepackage[table, dvipsnames]{xcolor}
\usepackage{xcolor,colortbl}
\usepackage{subfig}
\usepackage{diagbox}
\usepackage{url}
\usepackage{caption}
\usepackage{multirow}
\usepackage{mathptmx}       
\usepackage{helvet}         
\usepackage{courier}        
\usepackage{type1cm}        
\usepackage{makeidx}         
\usepackage{graphicx}        
\usepackage{multicol}        
\usepackage[bottom]{footmisc}

\makeindex             

\begin{document}

\title*{Automate Obstructive Sleep Apnea Diagnosis Using Convolutional Neural Networks}
\author{L. Feng and X. Wang}
\institute{Longlong Feng
\at Department of Mathematics, Wilfird Laurier University, Waterloo, ON, Canada \\ 
\email{feng0290@mylaurier.ca} \\
\\ Xu Wang 
\at Department of Mathematics \& MS2Discovery Interdisciplinary Research Institute \\
Wilfird Laurier University, Waterloo, ON, Canada \\ 
\email{xwang@wlu.ca}
}
\maketitle

\abstract*{Identifying the sleep problem severity from overnight polysomnography (PSG) recordings plays an important role in diagnosing and treating sleep disorders such as the Obstructive Sleep Apnea (OSA), which affects people especially children’s health. This analysis traditionally has been done by experts manually through visual inspections, which can be tedious, time-consuming, and is prone to subjective errors. Currently, there are many machine learning solutions used for analyzing and classifying PSG data. One of the proposed solutions is to use deep learning architectures such as Convolutional Neural Network (CNN) where the convolutional and pooling layers behave as feature extractors and some fully-connected (FCN) layers are used for making final predictions for the OSA severity classes. In this project, a CNN architecture with 1D convolutional layers and FCN layers for classification is proposed with the extensive tuning procedure of hyper-parameters. The PSG data for this project are from the Cleveland Children's Sleep and Health Study database and classification results confirm the effectiveness of the proposed CNN model. The proposed model of 1D CNN achieves excellent classification results without manually preprocesssing PSG signals such as feature extraction and feature reduction.}

\abstract{Identifying sleep problem severity from overnight polysomnography (PSG) recordings plays an important role in diagnosing and treating sleep disorders such as the Obstructive Sleep Apnea (OSA). This analysis traditionally is done by specialists manually through visual inspections, which can be tedious, time-consuming, and is prone to subjective errors. One of the solutions is to use Convolutional Neural Networks (CNN) where the convolutional and pooling layers behave as feature extractors and some fully-connected (FCN) layers are used for making final predictions for the OSA severity. In this paper, a CNN architecture with 1D convolutional and FCN layers for classification is presented. The PSG data for this project are from the Cleveland Children's Sleep and Health Study database and classification results confirm the effectiveness of the proposed CNN method. The proposed 1D CNN model achieves excellent classification results without manually preprocesssing PSG signals such as feature extraction and feature reduction.} 


\newpage
\section{Introduction and Background}
\label{sec:1}
When we sleep, our muscles relax. For the Obstructive Sleep Apnea (OSA) patients, the muscles in the back of throat can relax too much and collapse the airway, and lead to breathing difficulty. OSA presents with abnormal oxygenation, ventilation and sleep pattern. The prevalence of OSA has been reported to be between 1\% to 5\% \cite{OSA_children}. Children at risk need timely investigation and treatment.

The gold standard for diagnosing sleep disorders is polysomnography (PSG), which generates extensive data about biophysical changes during sleep. Studies of PSG assist doctors to diagnose sleep disorders and provide the baseline for an appropriate follow up. A clinical sleep study design based on PSG is to acquire several biological signals while patients are sleeping, These signals typically include electroencephalography (EEG) for monitoring brain activity, electromyogram (EMG) to measure muscle activity and Electrocardiography (ECG) for the electrical activity of heart over a period of sleep \cite{OSA_method}. 

In recent decades, various alternative methods have been proposed to minimize the number of biosignals required to detect and classify the OSA. These studies include traditional machine learning methods such as Support Vector Machine and linear discriminant analysis on signals such as ECG \cite{OSA_SVM}, respiratory signals \cite{OSA_respiratory}, a combination of extracted features and shallow neural network on heart rate variability and ECG derived respiration signal \cite{TRIPATHY_feature_based}. These studies focused on extracting time domain, frequency domain, and other nonlinear features from physiological signals and applying some feature selection techniques to reduce the number of dimensions comprising the feature space. However, this process can be labour-intensive, requires domain knowledge, and is particularly limited and costly for high-dimensional data. In addition, feature extraction is difficult for traditional machine learning techniques as the number of features increase dramatically.

Deep learning framework has proved its modeling ability in different PSG channels. McCloskey et al. employed a 2D-CNN model on spectrograms of nasal airflow signal, and their model achieved an average accuracy of 77.6\% on three severity levels \cite{CNN1D_lit_review}. Another more outstanding application of deep learning model came from the work of Cheng et al. in which researchers used a four layered Long Short Term Memory (LSTM) model on the RR-ECG signal and achieved an average accuracy of 97.80\% on the detection of OSA  \cite{LSTM_lit_review}. 

Though recurrent model (e.g., RNN, LSTM) can process time-series data and make sequential predictions, CNN can be trained to recognize the \textit{same} patterns (severity levels) on different subfields within fixed time windows. CNN saves time from manual scoring in the laboratory environment and makes the pre-screening stage easier in contrast to traditional methods. Moreover, in order to increase the model generalization ability, we tried to explore 1D-CNN models with different length of segmentations in EEG, ECG, EMG and respiratory channels. We focused on the model structure and utilized the fine-tuned model for pediatric OSA prediction in our study. 

The rest of this paper is organized as follows. Chapter 2 explains the data processing in detail. 
Chapter 3 displays the structure of the proposed 1D-CNN model. Evaluation and experimental results are presented in Chapter 4. Finally, Chapter 5 draws discussion and conclusion of the research.

\section{Cleveland Children's Sleep and Health Study Database}
\label{sec:2}
The data are retrieved from the National Sleep Research Resource (NSRR), which is a new National Heart, Lung, and Blood Institute resource designed to provide big data resources to the sleep research community. The PSG data are available from Cleveland Children's Sleep and Health Study (CCSHS) database. Each anonymous record includes a summary result of a 12-hour overnight sleep study (awake and sleep stages) including annotation files with scored events and PSG signals and being formatted as the European Data Format (EDF). 



The following channels are selected for the 1D CNN Modeling: 4 EEG channels (\textit{C3/C4} and \textit{A1/A2}), 3 EMG channels (\textit{EMG1, EMG2, EMG3}), 2 ECG channels (\textit{ECG1} and \textit{ECG2}), and 3 respiratory channels including airflow, thoracic and abdominal breathing.

\subsection{Individual Labeling}
\label{sec:3.1}

To define the target variable for this classification problem, each participant needs one label based on the OSA severity level. The Obstructive Apnea Hypopnea Index (oahi3) is used to indicate the severity of sleep apnea. It is represented by the number of apnea and hypopnea events per hour of sleep. It combines AHI and oxygen desaturation to give an overall sleep apnea severity score that evaluates both the number of sleep disruptions and the degree of oxygen desaturation (low oxygen level in the blood). The values of \textit{oahi3} are used as the thresholds for grouping the participants. The number of participants with different severity levels are shown in Table \ref{tab:labeling}.

\setlength{\tabcolsep}{7pt}
\renewcommand{\arraystretch}{1.2}
\begin{table}[hbt!]
\centering
 \begin{tabular}{||c | c | c ||} 
 \hline
 Obstructive Apnea Hypopnea Index & Level of Severity  & Number of Participants\\
 \hline
0 $<$ oahi3 $\leq$ 1 & NL (Normal) & 362\\ 
 \hline
1 $<$ oahi3 $\leq$ 5 & MIN (Minor) & 139\\
 \hline
5 $<$ oahi3 $\leq$ 10 & MOD (Moderate) & 8\\
 \hline
10 $<$ oahi3 & SV (Severe) & 8\\ 
 \hline
\end{tabular}
\captionsetup{justification=centering}
\caption{Grouping participants using oahi3 values} \label{tab:labeling}
\end{table}

The dataset has an imbalanced response variable (362 normal / 139 minor / 8 moderate / 8 severe). Those minority classes (moderate and severe) are our most interest. We tried to train classifier to learn more from moderate and severe level data. Under-sampling method was applied during the data pre-processing stage, i.e., we randomly selected an equal number of samples (i.e., 8 participants) from each of the normal and minor groups. Overall, there are 32 participants in the final study data set. In this project, we conduct data pre-processing and CNN modeling on the data in EDF format which have a total size of 13 GB.

\subsection{Data Preprocessing}
\label{sec:3.2}
\begin{figure}[hbt!]
    \centering
    \includegraphics[width=.9\textwidth]{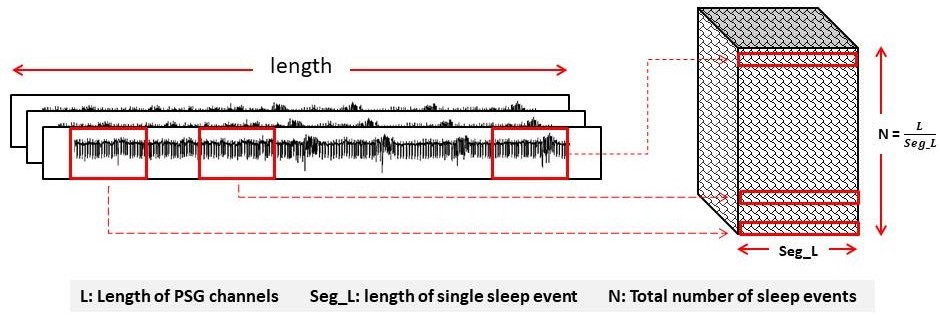}\\
    \caption{Demonstration of Channel Division}
    \label{fig:segmentation}
\end{figure}

This experiment focuses on the sleep data. The beginning and ending \textit{awake} signals could be treated as noise and need to be removed. Secondly, the deep learning algorithms tend to be difficult to train when the length of time series is very long. Figure \ref{fig:segmentation} presents a segmentation strategy, i.e., dividing the time series into smaller chunks. 

each segment was labeled as the same severity level as the participant. In other words, the segments would inherit the severity label from the participant they belong to. With a starting length of L time steps, one channel is divided into blocks of sequence \emph{Seq\_L} yielding about \emph{L / Seq\_L} of new events (or rows) of shorter length (\emph{N}).

The PSG data were segmented into 1-minute long events. For the ECG channel  (frequency of 256) a 1-minute event has a length of 15360 (256$\times$60) data points. An individual has a 8.24-hour ECG channel, which would have 1D time series data with length of 7595520. After segmentation, the long series data turned into a tensor with dimension 494$\times$15360, which indicates 494 events (a length of 15360 for each). Since we have 32 selected participants and 2 ECG channels for each participant, the input tensor has the dimension of 15824 (\emph{N})$\times$15360 (\emph{Seq\_L})$\times$2 (channels).

With the data segmentation, the length of each time-series is shorter and will be helpful in model training; and the number of data points has increased by a factor of \emph{L / Seq\_L} (number of instances or rows) providing a larger data set to train on. 


Since different channels (e.g., ECG, EMG) were measured in different amplitudes, therefore, the last step of data processing is to normalize the PSG data with zero mean and unit standard deviation.



\section{1D-CNN Architecture}
\label{sec:3}
The convolutional layer and max-pooling layer play the key roles in the CNN’s feature extraction mechanism. The output of convolutional layer of the $\ell^{th}$ layer can be calculated as in Formula \ref{eq:convolutional_layer_eq}:

\begin{eqnarray}
    C_{k}^{(\ell)} = ReLU (\sum_{c}W_{k}^{(\ell), c} \ast X^{(\ell-1), c} + B_{k}^{\ell}),
 \label{eq:convolutional_layer_eq}
\end{eqnarray}

where $k$ represents the filter number, $c$ denotes the channel number of the input $X^{\ell -1}$, $W_{k}^{(\ell), c}$ is the $k^{th}$ convolutional filter to the $c^{th}$ channel, and $ B_{k}^{\ell}$ is the bias to the $k^{th}$ filter, and $\ast$ is the dot product operation.

The max-pooling layer is a sub-sampling function selecting the maximum value within a fixed size filter. After the convolution-pooling blocks, one fully connected layer of neurons which have full connections to all activations in the previous layer, as in the regular Neural Networks. At the end of the convolutional layers, the data were flattened and passed onto the Dropout layer before the softmax classifier. 

\begin{figure}[hbt!]
    \centering
    \includegraphics[width=.66\textwidth]{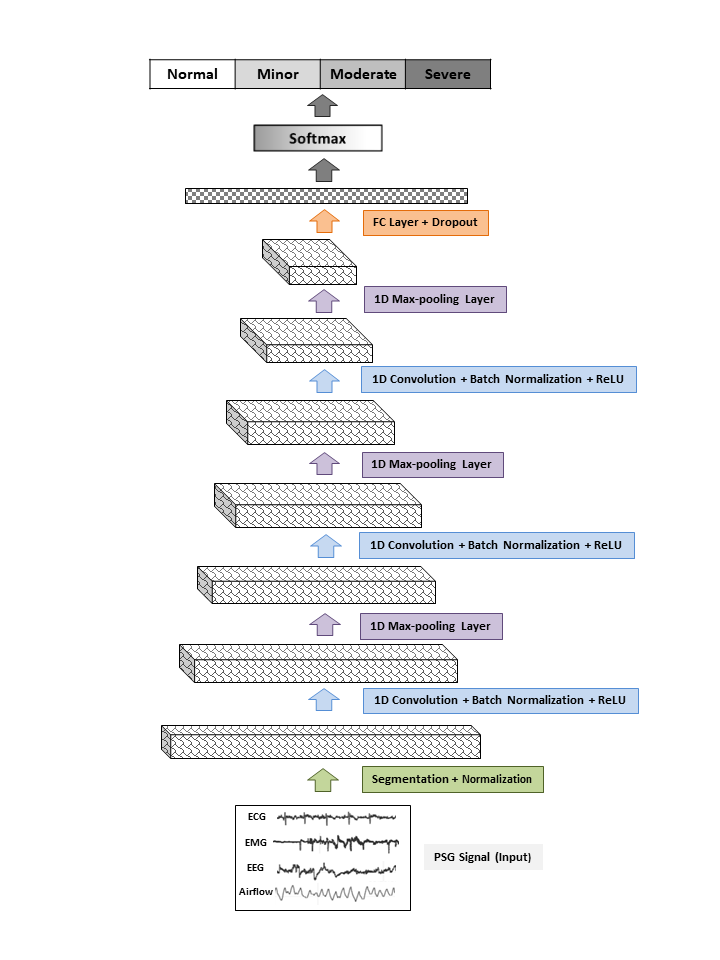}\\
    \caption{The Proposed 1D-CNN Architecture}
    \label{fig:1dCNN}
\end{figure}

Figure \ref{fig:1dCNN} shows the structure of the 1D CNN model proposed in this project. It contains 3 convolutional and 3 max-pooling layers. We focused our efforts on the CNN building and began the investigation of the CNN method initially by performing a grid search of several hyperparameters. 

For each participant, his or her PSG data were served for either training or test data, not for both. We implemented a two-level stratified random sampling. In details, there were 2 splitting steps among 32 participants: firstly, 8 were randomly selected as test participants (i.e., 2 participants were randomly selected for each severity level); secondly, the remaining 24 participants were split into two groups: 18 participants for training set and 6 participants for validation set. The tensorflow graph was fed with batches of the training data and the hyperparameters were tuned on a validation set. Finally the trained model was evaluated on the test set. 

The CNN model was trained in a fully supervised manner, and the gradients were back-propagated from the softmax layer to the convolutional layers. The network parameters were optimized by minimizing the cross-entropy loss function based on the gradient descent with the Adam updating rule and a learning rate of 0.0001.
\setlength{\tabcolsep}{3pt}
\renewcommand{\arraystretch}{1.1}
\begin{table}[hbt!]
\centering
 \begin{tabular}{||c | c c c c c||} 
 \hline
  CNN Layer & $\#$ of filters & Filter Size & Stride & Padding & Activation Function \\ 
 \hline\hline
 Conv 1 & 46 & 10 & 2 & No & Relu \\ 
 \hline
 Pooling 1 & -- & 10 & 2 & No & -- \\
 \hline
 Conv 2 & 92 & 10 & 2 & No & Relu \\
 \hline
 Pooling 2 & -- & 10 & 2 & No & -- \\
 \hline
 Conv 3 & 184 & 20 & 2 & No & Relu \\
 \hline
 Pooling 3 & -- & 20 & 5 & No & -- \\
 \hline
\end{tabular}
\caption{CNN model structure with optimal parameters} \label{tab:CNN_hyperparameters}
\end{table}

Table \ref{tab:CNN_hyperparameters} presents the final values of parameters within each layer. Dropout rate of $0.5$ was used as it is the general setting for CNN models. Model classification performance is evaluated by using the following metrics: classification accuracy, cross-entropy loss, precision, recall and F1-score. While accuracy and loss can be used for evaluating the overall performance, some other metrics can be used to measure the performance of specific class.

\section{Results and Analysis} 
\label{sec:4}
\begin{figure}[hbt!]
    \centering
    \includegraphics[width=.95\textwidth]{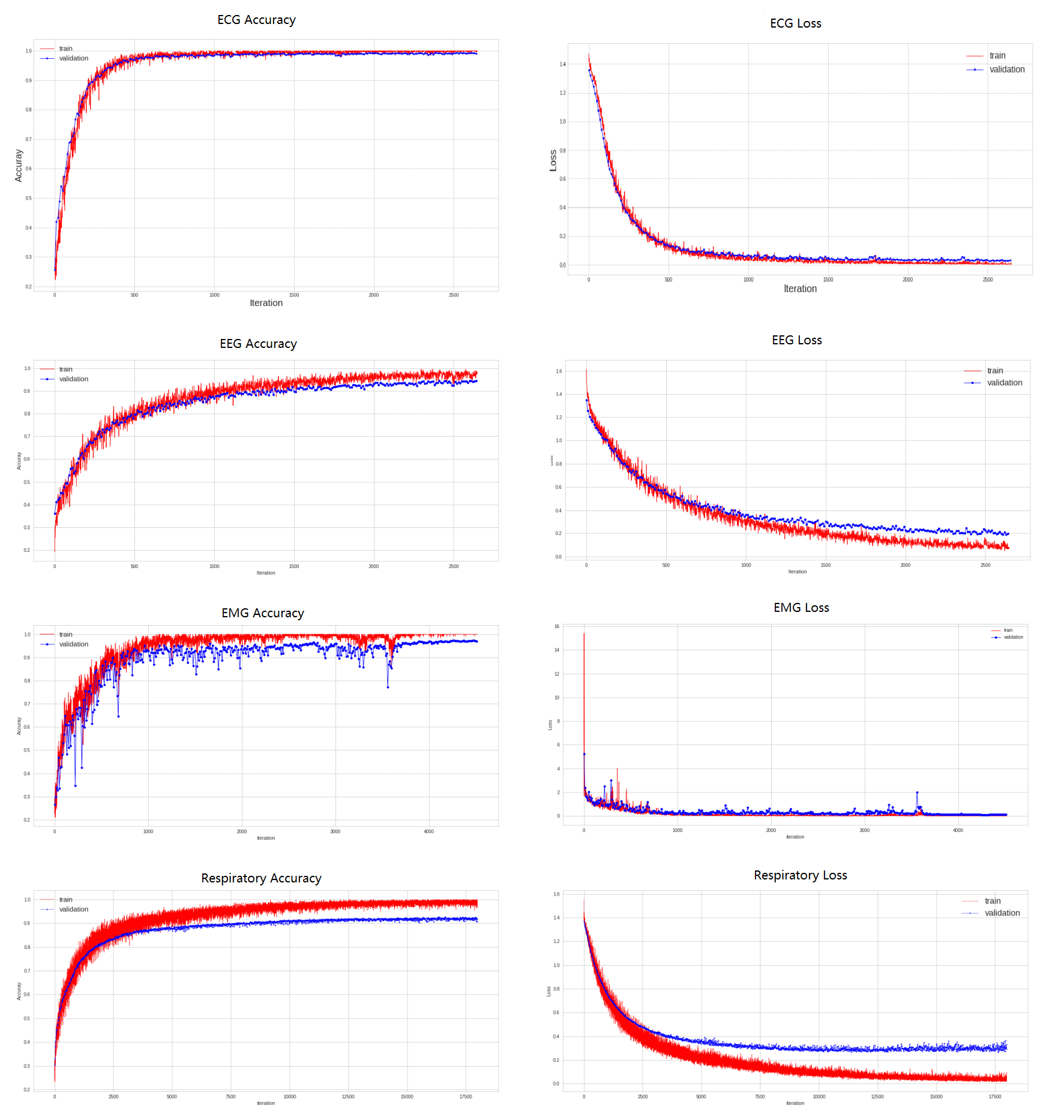}
    \caption{Accuracy and Loss of the Proposed CNN model for OSA Detection}
    \label{fig:training}
\end{figure}

Figure \ref{fig:training} shows the learning curve on training and validation phases. Accuracy and loss were obtained with various number of iterations. The accuracy increases as the number of iteration increases, and the loss decreases at the same time. The accuracy and the loss reach stable values after iterative learning on both phases. 

For ECG, we can observe the stable accuracy and loss values after 1000 iterations (Training acc: 0.9987, loss: 0.0114; Validation acc: 0.9916, loss: 0.0289). For EEG, the accuracy and the loss start to converge to a value after 2500 iterations (Training acc: 0.9718, loss: 0.0945; Validation acc: 0.9447, loss: 0.1985). For EMG, the accuracy and the loss become stable after 4000 iterations (Training acc: 0.9999, loss: 0.0013; Validation acc: 0.9707, loss: 0.1131). However, there are a large number of big fluctuations before the convergence during the learning process. This means some portion of the randomness: (1) The Dropout method could cause the network to keep only some portion of neurons (weights) on each iteration. Sometimes those neurons do not fit the current batch well, and this may cause large fluctuations; (2) There is randomness in initialization and data sampling for SGD in back-propagation. 

For Respiratory, we can see the train and validation accuracy begin to stay steady with similar values indicating slight overfitting in the classification (Training acc: 0.9854, loss: 0.0378; Validation acc: 0.9180, loss: 0.2945). 

\setlength{\tabcolsep}{7pt}
\renewcommand{\arraystretch}{1.3}
\begin{table}[hbt!]
\centering
 \scriptsize
 \captionsetup{font=normal}
 \begin{tabular}{|| c | c | c  c | *{4}{>{\columncolor{blue!20}}c} ||} 
 \hline
 \rowcolor{white}
 Channels(\#) & Dataset & Accuracy & Loss & Class & Precision & Recall & F1-Score \\
 \hline\hline
 \multirow{8}{*}{ECG (2)} & \multirow{4}{*}{Training} & \multirow{4}{*}{0.9987} & \multirow{4}{*}{0.0114} & NL & 0.9997 & 0.9997 & 0.9994\\ 
                        & & & & \cellcolor{white} MIN & \cellcolor{white} 0.9980 & \cellcolor{white} 0.998 & \cellcolor{white} 0.9980\\ 
                        & & & & MOD & 0.9982 & 0.9982 & 0.9982\\
                        & & & & \cellcolor{white} SV & \cellcolor{white} 0.9988 & \cellcolor{white} 0.9994 & \cellcolor{white} 0.9991 \\\cline{2-8}
                        
                        & \multirow{4}{*}{Test} & \multirow{4}{*}{0.9897} & \multirow{4}{*}{0.0289} & 
                        NL & 0.9862 & 0.9921 & 0.9891 \\
                        & & & & \cellcolor{white} MIN & \cellcolor{white} 0.9990 & \cellcolor{white} 0.9773 & \cellcolor{white} 0.9880 \\
                        & & & & MOD & 0.9894 & 0.9961 & 0.9927 \\
                        & & & & \cellcolor{white} SV & \cellcolor{white} 0.9843 & \cellcolor{white} 0.9940 & \cellcolor{white} 0.9891 \\
 \hline
 
  \hline\hline
 \multirow{8}{*}{EEG (4)} & \multirow{4}{*}{Training} & \multirow{4}{*}{0.9718} & \multirow{4}{*}{0.0945} & NL & 0.9753 & 0.9741 & 0.9747 \\ 
                        & & & & \cellcolor{white} MIN & \cellcolor{white} 0.9784 & \cellcolor{white} 0.9820 & \cellcolor{white} 0.9802 \\ 
                        & & & & MOD & 0.9721 & 0.9684 & 0.9703 \\
                        & & & & \cellcolor{white} SV & \cellcolor{white} 0.9609 & \cellcolor{white} 0.9621 & \cellcolor{white} 0.9615 \\\cline{2-8}
                        
                        & \multirow{4}{*}{Test} & \multirow{4}{*}{0.9463} & \multirow{4}{*}{0.1985} & NL & 0.9394 & 0.9587 & 0.9490 \\
                        & & & & \cellcolor{white} MIN & \cellcolor{white} 0.9415 & \cellcolor{white} 0.9741 & \cellcolor{white} 0.9575 \\
                        & & & & MOD & 0.9682 & 0.9166 & 0.9417 \\
                        & & & & \cellcolor{white} SV & \cellcolor{white} 0.9373 & \cellcolor{white} 0.9354 & \cellcolor{white} 0.9363 \\
 \hline
 
   \hline\hline
 \multirow{8}{*}{EMG (3)} & \multirow{4}{*}{Training} & \multirow{4}{*}{0.9999} & \multirow{4}{*}{0.0013} & NL & 1.0000 & 1.0000 & 1.0000 \\ 
                        & & & & \cellcolor{white} MIN & \cellcolor{white} 1.0000 & \cellcolor{white} 0.9997 & \cellcolor{white} 0.9999 \\ 
                        & & & & MOD & 0.9997 & 1.0000 & 0.9999 \\
                        & & & & \cellcolor{white} SV & \cellcolor{white} 1.0000 & \cellcolor{white} 1.0000 & \cellcolor{white} 1.0000 \\\cline{2-8}
                        
                        & \multirow{4}{*}{Test} & \multirow{4}{*}{0.9581} & \multirow{4}{*}{0.1132} & NL & 0.9518 & 0.9312 & 0.9414 \\
                        & & & & \cellcolor{white} MIN & \cellcolor{white} 0.9660 & \cellcolor{white} 0.9601 & \cellcolor{white} 0.9631 \\
                        & & & & MOD & 0.9823 & 0.9712 & 0.9767 \\
                        & & & & \cellcolor{white} SV & \cellcolor{white} 0.9329 & \cellcolor{white} 0.9696 & \cellcolor{white} 0.9509 \\
 \hline
 
  \hline\hline
 \multirow{8}{*}{Respiratory (3)} & \multirow{4}{*}{Training} & \multirow{4}{*}{0.9854} & \multirow{4}{*}{0.0378} & NL & 0.9857 & 0.9857 & 0.9857 \\ 
                        & & & & \cellcolor{white} MIN & \cellcolor{white} 0.9834 & \cellcolor{white} 0.9849 & \cellcolor{white} 0.9842 \\ 
                        & & & & MOD & 0.9895 & 0.9880 & 0.9888 \\
                        & & & & \cellcolor{white} SV & \cellcolor{white} 0.9828 & \cellcolor{white} 0.9828 & \cellcolor{white} 0.9828 \\\cline{2-8}
                        
                        & \multirow{4}{*}{Test} & \multirow{4}{*}{0.9199} & \multirow{4}{*}{0.2945} & NL & 0.9147 & 0.9147 & 0.9147 \\
                        & & & & \cellcolor{white} MIN & \cellcolor{white} 0.9447 & \cellcolor{white} 0.9053 & \cellcolor{white} 0.9246 \\
                        & & & & MOD & 0.9323 & 0.9194 & 0.9258 \\
                        & & & & \cellcolor{white} SV & \cellcolor{white} 0.8899 & \cellcolor{white} 0.9408 & \cellcolor{white} 0.9147 \\
 \hline
\end{tabular}
\vspace{2mm}
\caption{The CNN Evaluation Metrics} \label{tab:CNN_Evaluation_Metrics}
\end{table}

\setlength{\tabcolsep}{7pt}
\renewcommand{\arraystretch}{1.1}
\begin{table}[hbt!]
    \centering
    \scriptsize

    \begin{tabular}{|| c | c c c c | c c c c ||}
        \cline{2-9}

        \multicolumn{1}{c|}{} & & \multicolumn{2}{c}{ECG Training} & & & \multicolumn{2}{c}{ECG Test} & \\ \hline
        \rowcolor{blue!20}
        \diagbox[]{True}{Predict} & NL & MIN & MOD & SV & NL & MIN & MOD & SV\\\hline
        \cellcolor{blue!20} NL & 3321 & 0 & 1 & 2 & 1000 & 0 & 3 & 5\\
        \cellcolor{blue!20} MIN & 0 & 3511 & 5 & 2 & 10 & 1032 & 6 & 8 \\
        \cellcolor{blue!20} MOD & 1 & 5 & 3378 & 0 & 1 & 0 & 1022 & 3 \\
        \cellcolor{blue!20} SV & 0 & 2 & 0 & 3340 & 3 & 1 & 2 & 1000 \\
        \hline

        \multicolumn{1}{c|}{} & & \multicolumn{2}{c}{EEG Training} & & & \multicolumn{2}{c}{EEG Test} & \\ \hline
        \rowcolor{blue!20}
        \diagbox[]{True}{Predict} & NL & MIN & MOD & SV & NL & MIN & MOD & SV\\\hline
        \cellcolor{blue!20} NL & 3239 & 14 & 13 & 59 & 976 & 12 & 4 & 26\\
        \cellcolor{blue!20} MIN & 9 & 3445 & 34 & 20 & 7 & 1014 & 11 & 9\\
        \cellcolor{blue!20} MOD & 15 & 40 & 3279 & 52 & 28 & 30 & 945 & 28\\
        \cellcolor{blue!20} SV & 58 & 22 & 47 & 3222 & 28 & 21 & 16 & 941\\
        \hline

        \multicolumn{1}{c|}{} & & \multicolumn{2}{c}{EMG Training} & & & \multicolumn{2}{c}{EMG Test} & \\ \hline
        \rowcolor{blue!20}
        \diagbox[]{True}{Predict} & NL & MIN & MOD & SV & NL & MIN & MOD & SV\\\hline
        \cellcolor{blue!20} NL & 3546 & 0 & 0 & 0 & 731 & 14 & 9 & 31\\
        \cellcolor{blue!20} MIN & 0 & 3745 & 1 & 0 & 11 & 795 & 5 & 17\\
        \cellcolor{blue!20} MOD & 0 & 0 & 3601 & 0 & 9 & 7 & 775 & 7\\
        \cellcolor{blue!20} SV & 0 & 0 & 0 & 3571 & 17 & 7 & 0 & 765\\
        \hline

        \multicolumn{1}{c|}{} & & \multicolumn{2}{c}{Respiratory Training} & & & \multicolumn{2}{c}{Respiratory Test} & \\ \hline
        \rowcolor{blue!20}
        \diagbox[]{True}{Predict} & NL & MIN & MOD & SV & NL & MIN & MOD & SV\\\hline
        \cellcolor{blue!20} NL & 3714 & 18 & 14 & 22 & 461 & 10 & 14 & 19\\
        \cellcolor{blue!20} MIN & 16 & 3912 & 13 & 31 & 10 & 478 & 14 & 26\\
        \cellcolor{blue!20} MOD & 17 & 17 & 3786 & 12 & 20 & 7 & 468 & 14\\
        \cellcolor{blue!20} SV & 21 & 31 & 13 & 3723 & 13 & 11 & 6 & 477\\
        \hline
        
    \end{tabular}\hfil 
    \vspace{2mm}
    \caption{Confusion matrices from the CNN model on training and test data}
    \label{tab:Confusion_matrix_final}
\end{table} 

\newpage
The evaluation metrics and confusion matrices for all channels with training and test data are presented in Tables \ref{tab:CNN_Evaluation_Metrics} and \ref{tab:Confusion_matrix_final} respectively. The results from Table \ref{tab:Confusion_matrix_final} are summarized in Table \ref{tab:CNN_Evaluation_Metrics}. It can be observed from Table \ref{tab:CNN_Evaluation_Metrics} that, for the test data, the CNN model can achieve 98.97\% for ECG, 94.63\% for EEG, 95.81\% for EMG, and 91.99\% for Respiratory; We can also verify the training curves from Figure \ref{fig:training} by checking the training accuracy score from Table \ref{tab:CNN_Evaluation_Metrics} and the classified results from Table \ref{tab:Confusion_matrix_final}. Furthermore, the precision, recall and F1-score for each class are collected in Table \ref{tab:CNN_Evaluation_Metrics}. 

For ECG, the model can achieve a value of $> 99\%$ for all three metrics for all classes on the training data and $>97\%$ for the test data; For EEG, the model achieves $>96\%$ score for training data, and $>91\%$ for the test data. 

For EMG, the scores of 1.0000 are obtained in the training phase on all classes, which means the perfect classification for the training data during the learning process, while the scores of $>93.29\%$ are obtained from the test data. 

Similarly, for Respiratory, CNN achieves scores of $>98\%$ for the training and slightly lower scores, which are over $>88.99\%$ for the test data. The reason why there exists the gap between training and test scores can be that the respiratory signal sensors is different from ECG, EEG and EMG. In this case, the signal in the respiratory system may not be sensitive enough to detect small changes when OSA happens. Table \ref{tab:Confusion_matrix_final} displays the classification details on the training and test data.  

\section{Conclusion and Discussion} %
\label{sec:5}
Firstly, with the correct hyper-parameter setup, our 1D-CNN model can successfully extract the temporal features from the PSG data and achieve high performance in OSA detection for different channels; secondly, our well trained CNN model can be an efficient tool for clinicians to identify OSA severity without manually going through tons of PSG data. Furthermore, our CNN models can replace the traditional data processing such as signal extraction and transforming, which can be time-consuming and labour-intense.

There are some limitations of our work. Firstly, only a small sample of 32 subjects was investigated in this study. Secondly, we used ECG, EEG, EMG and Respiratory channels to build CNN models separately, so there was no cross-checking between different channels. Lastly, our CNN model is slow to be trained without GPU. The well-trained models require a big data set and the fine-tuned hyperparameters in the training step. 

The future work can aim at feeding the four single CNN models into an ensemble-like model to making a prediction. There are other possible architectures that would be of great interest for this problem. One of most popular deep learning architectures that models sequence and time-series data is the long-short-term memory (LSTM) cells within recurrent neural networks (RNN). 

\begin{acknowledgement}
We are so grateful that National Sleep Research Resource (NSRR) allows us to use the PSG data from Cleveland Children's Sleep and Health Study. The project is supported by Natural Sciences and Engineering Research Council of Canada (NSERC).
\end{acknowledgement}

\bibliographystyle{unsrt}
\bibliography{referenc}

\begin{thebibliography}{1}

\bibitem{OSA_children}
Eleonora Dehlink and Hui-Leng Tan.
\newblock Update on paediatric obstructive sleep apnoea.
\newblock {\em Journal of Thoracic Disease}, 8(2), 2016.

\bibitem{OSA_method}
M.~K. {Moridani}, M.~{Heydar}, and S.~S. {Jabbari Behnam}.
\newblock A reliable algorithm based on combination of emg, ecg and eeg signals
  for sleep apnea detection : (a reliable algorithm for sleep apnea detection).
\newblock In {\em 2019 5th Conference on Knowledge Based Engineering and
  Innovation (KBEI)}, pages 256--262, Feb 2019.

\bibitem{OSA_SVM}
W.~S. {Almuhammadi}, K.~A.~I. {Aboalayon}, and M.~{Faezipour}.
\newblock Efficient obstructive sleep apnea classification based on eeg
  signals.
\newblock In {\em 2015 Long Island Systems, Applications and Technology}, pages
  1--6, May 2015.

\bibitem{OSA_respiratory}
C.~{Varon}, A.~{Caicedo}, D.~{Testelmans}, B.~{Buyse}, and S.~{Van Huffel}.
\newblock A novel algorithm for the automatic detection of sleep apnea from
  single-lead ecg.
\newblock {\em IEEE Transactions on Biomedical Engineering}, 62(9):2269--2278,
  Sep. 2015.

\bibitem{TRIPATHY_feature_based}
R.K. Tripathy.
\newblock Application of intrinsic band function technique for automated
  detection of sleep apnea using hrv and edr signals.
\newblock {\em Biocybernetics and Biomedical Engineering}, 38(1):136 -- 144,
  2018.

\bibitem{CNN1D_lit_review}
Stephen McCloskey, Rim Haidar, Irena Koprinska, and Bryn Jeffries.
\newblock {\em Detecting Hypopnea and Obstructive Apnea Events Using
  Convolutional Neural Networks on Wavelet Spectrograms of Nasal Airflow},
  pages 361--372.
\newblock 06 2018.

\bibitem{LSTM_lit_review}
Maowei Cheng, Worku Sori, Feng Jiang, Adil Khan, and Shaohui Liu.
\newblock Recurrent neural network based classification of ecg signal features
  for obstruction of sleep apnea detection.
\newblock pages 199--202, 07 2017.

\end{thebibliography}
\end{document}